\documentclass[10pt,twocolumn,letterpaper]{article}

\usepackage[pagenumbers]{wacv} % To force page numbers, e.g. for an arXiv version

\usepackage{comment}
\usepackage{makecell}
\usepackage{multirow}

\definecolor{wacvblue}{rgb}{0.21,0.49,0.74}
\usepackage[pagebackref,breaklinks,colorlinks,allcolors=wacvblue]{hyperref}

\def\wacvPaperID{2319} % *** Enter the WACV Paper ID here
\def\confName{WACV}
\def\confYear{2027}

\title{CoViG: Coarsened Vision GNN for Image Classification}

\author{Martial Guidez, Stefan Duffner, Christophe Garcia\\
INSA Lyon, CNRS, Université Claude Bernard Lyon 1, LIRIS, UMR5205, 69621 Villeurbanne, France\\
{\tt\small martial.guidez@gmail.com}
}

\begin{document}
\maketitle

\begin{abstract}
Although Graph Neural Networks are widely used in various domains, they remained mostly unexplored in computer vision. Convolutional Neural Networks and Vision Transformers were leading the field, until the quite recent introduction of Vision Graphs (ViG) that treat images as a set of nodes and apply graph convolution on their k-NN graphs. ViGs show impressive accuracy on image classification but their computational cost remains relatively high. Inspired by optimization methods for GNNs, we present CoViG, a novel approach combining graph coarsening and ViGs. 
Our method introduces an additional MinCUT pooling step within the Grapher component of intermediate ViG blocks. 
By simplifying the intermediate graphs and thus decreasing the number of nodes of specific blocks, our approach considerably reduces the computational cost of ViG during inference.
We evaluated our approach on ImageNet with several ViG architectures, and we obtained accuracies equivalent to the original ViG models with only about 51\% of FLOPs.
\end{abstract}

\section{Introduction}
\label{sec:intro}

Although Convolutional Neural Networks (CNNs) and Vision Transformer (ViTs) are the predominant neural network architectures of state-of-the-art computer vision models, a new type of architecture called Vision Graphs (ViGs) has been introduced recently~\cite{Han2022VisionGA} showing excellent performances on the standard ImageNet image classification benchmark.
Similar to ViT, ViGs divide the input image into non-overlapping patches, and then compute embeddings that are processed further by the backbone. 
But, instead of the Transformer blocks that implicitly integrate the contextual information using attention, ViGs model these spatial relations and visual similarities more explicitly using graphs encoded with a Graph Neural Network (GNN) based architecture. 

Although the costly attention mechanism is avoided in ViGs, computational complexity remains still relatively high due to the graph generation process and GCN at each block of the model.
Much effort has been spent in the last decade to increase the efficiency of CNNs and ViTs (and Transformers in general), both in terms of computation and memory. 
However, only a few recent works have specifically tackled the reduction of ViG complexity~\cite{Munir2023MobileViGGS, Munir2024GreedyViGDA, Parikh2025ClusterViGEG}.
The main computational bottleneck is the k-Nearest-Neighbor-based graph construction in each ViG block, where the similarity between each pair of patch embeddings needs to be computed, and a Graph Convolutional Network (GCN) is applied on this graph. 

\begin{figure}[t]
    \centering
    \includegraphics[width=\linewidth]{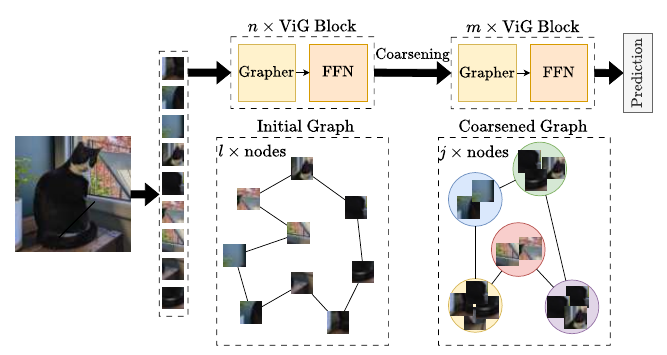}
    \caption{Overview of the CoViG architecture: graph coarsening reduces the node set from $l$ nodes to $j$ super-nodes between two stages of ViG blocks. In this illustration, only a single coarsening step is applied.}
    \label{fig:covig}
\end{figure}

Our proposed method directly addresses this by introducing a graph coarsening approach at pre-determined stages (blocks) of the model.
More specifically, we replace one or several of the Grapher (GN) components with, what we call a Coarsened Grapher that simplifies the graph into a reduced version with much fewer nodes. This process is represented in \autoref{fig:covig}. This reduction then propagates to all subsequent blocks of the model and has a direct impact on the computational efficiency. 
Our coarsening approach is based on the MinCUT Pooling algorithm~\cite{Bianchi2019MincutPI} that learns the clustering jointly with the rest of the model and a task-specific loss, as opposed to other GNN or ViG optimisation methods where this clustering is performed independently~\cite{Dhillon2007WeightedGC, Parikh2025ClusterViGEG}. 

Our contributions can be summarised as follows:
\begin{itemize}
% CoViG framework
\item We propose a new Vision Graph-based architecture with reduced computational complexity thanks to an integrated and differentiable graph coarsening approach based on MinCUT Pooling.
% Flexible module
\item By introducing a new Coarsened Grapher module, the proposed approach is flexible and adaptable to different (pre-trained) ViG models. Before training or fine-tuning, one simply has to choose one or several blocks that should be reduced and determine the desired compression rate for each coarsening step.
% Thorough evaluation
\item We extensively evaluated our approach on the ImageNet benchmark, with different coarsening configurations and rates, and with different ViG models, and we were able to reduce the FLOPs up to a factor of 2 without decreasing classification accuracy.
\end{itemize}

\section{Related Work}
\label{sec:sota}

This section provides a brief overview of the different architectures used for image classification and the presentation of the compression methods that are applied to them. 

\subsection{CNN and ViT}

% CNN
Convolutional Neural Networks (CNNs) have been the dominant paradigm in computer vision, driving remarkable advances in image classification \cite{Krizhevsky2012ImageNetCW}, object detection \cite{Ren2015FasterRT}, and semantic segmentation \cite{Shelhamer2014FullyCN}. LeNet \cite{LeCun1998GradientbasedLA} established the fundamental principles of local receptive fields, weight sharing, and hierarchical feature extraction. It was followed by AlexNet \cite{Krizhevsky2012ImageNetCW}, a breakthrough on the ImageNet Large Scale Visual Recognition Challenge (ILSVRC). This work paved the way for the well known VGGNet \cite{Simonyan2014VeryDC}, ResNet \cite{He2015DeepRL}, and MobileNet \cite{Howard2017MobileNetsEC}. 

% CNN compression
The generalized use of CNNs led to the need of compressing the networks to alleviate their memory footprint and computational complexity \cite{Cheng2017ASO, Han2021DynamicNN}. Many methods have been developed such as Pruning \cite{Han2015LearningBW, Srinivas2015DatafreePP}, Knowledge Distillation \cite{Ba2013DoDN,Hinton2015DistillingTK, Romero2014FitNetsHF}, Early-Exit \cite{Park2015BiglittleDN, Teerapittayanon2016BranchyNetFI}, and Quantization \cite{Gong2014CompressingDC, Gupta2015DeepLW}.
CNNs have long dominated the field of computer vision, primarily due to their ability to capture spatial hierarchies through convolutional layers compared to classic non-convolutional Deep Neural Networks (DNN).

% Vision Transformer
After the rise of attention in the field of NLP \cite{Vaswani2017AttentionIA}, Dosovitsky et al. \cite{Dosovitskiy2020AnII} have developed ViT for image recognition.
Compared to CNNs, ViTs capture global image context more effectively especially where long-range dependencies are crucial. However, it usually needs to be trained on very large-scale datasets. ViTs are also computationally expensive due to the self-attention mechanism, which scales quadratically with the number of image patches, leading to high memory and computation resource requirements. 
These computational costs have motivated the research of compression techniques for ViT. 

% CNN compression on ViT
Some CNN compression techniques have been successfully applied to ViT. DeiT \cite{Touvron2020TrainingDI} have applied knowledge distillation to ViT by adding a specific token. Liu \etal \cite{Liu2021PostTrainingQF} also achieved post-training quantization on ViT.

% ViT architecture optimization 
Other methods also try to reduce the complexity of the self-attention mechanism, which scales quadratically with the patch number. For example, Wang \etal proposed pyramid architecture \cite{Wang2021PyramidVT} to improve ViT performances while reducing the network cost. Liu \etal proposed the Swin Transformer \cite{Liu2021SwinTH}, which has a linear computational complexity with respect to the image size. 

% ViT specific compression
Finally, some compression methods leverage ViT-specific architecture. For example, Rao \etal introduce DynamicViT \cite{Rao2021DynamicViTEV}, to prune unnecessary tokens. Bolya \etal explore token merging in ToMe \cite{Bolya2022TokenMY}.

\subsection{From GNN to Vision GNN}

%GNN
In many applications, objects and relationships between examples in a dataset can be represented as a graph. However, common machine learning algorithms rely on the assumption that instances are independent of each other. This is invalid for graph data. 
This leads to ever-increasing use of graph-based methods in different fields.
Neural networks that operate on graphs (GNN) have been introduced by Gori \etal \cite{Gori2005ANM}.
Convolution-like propagation in GNN was first introduced by Duvenaud \etal \cite{Duvenaud2015ConvolutionalNO} and later improved to became the modern GCN by Micheli \cite{Micheli2009NeuralNF}.
%
% GCN CV applications
GCN have been used widely in computer vision applications and specific visual tasks with naturally constructed graphs, such as point cloud classification and segmentation \cite{Wang2018DynamicGC,Landrieu2017LargeScalePC}, scene graph generation \cite{Xu2017SceneGG}, and action recognition \cite{Yan2018SpatialTG}.

\begin{figure*}[t]
    \begin{subfigure}{.4\textwidth}
        \includegraphics[width=\linewidth]{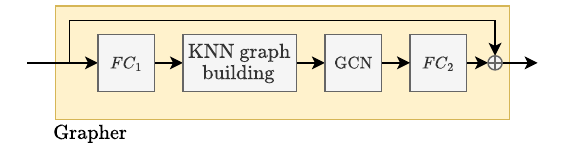}
        \caption{}
        \label{fig:block}
    \end{subfigure} 
    \hfill
    \begin{subfigure}{.55\textwidth}
        \includegraphics[width=\linewidth]{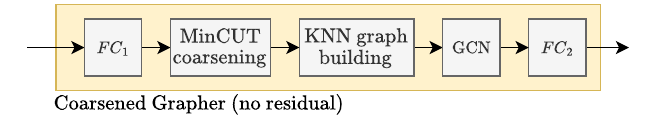}
        \caption{}
        \label{fig:coarse_block}
    \end{subfigure}
    \caption{Comparison of the baseline Grapher block (a) and the proposed coarsened Grapher block (b).}
\end{figure*}

%ViG
The idea behind Vision Graphs (ViG) \cite{Han2022VisionGA} is to model the semantic spatial relationship of image parts (i.e.~patches) together with their content in a graph.
This approach introduces a GCN-based backbone that directly processes image data. It uses patches as nodes and builds a graph with the top first-order similarity as neighbors. 
This model outperformed state-of-the-art CNN and ViT-based architectures on standard image classification benchmarks. They proposed two versions of their network: an isotropic version and a pyramid version (PViG).
%MobileViG
Inspired  by this work, Munir \etal have proposed an hybrid CNN-GNN architecture MobileViG \cite{Munir2023MobileViGGS} based on PViG, that uses Sparse Vision Graph Attention instead of the previously used k-Nearest Neighbors (k-NN) graph attention. 
%GreedyViG
Munir \etal also proposed GreedyViG \cite{Munir2024GreedyViGDA} which uses Dynamic Axial Graph to replace k-NN graph construction on PViG. 
%ClusterViG
Parikh \etal have proposed ClusterViG \cite{Parikh2025ClusterViGEG}, introducing Dynamic Efficient Graph Convolution. It reduces the computational expense of image graph construction by partitioning the image patch feature grid into several partitions and performing graph construction in parallel within each partition. K-NN is performed locally within each partition in parallel, reducing the cost from $O\left(N^2\right)$ to $O\left(N^2/k^2\right)$. The number of nodes $N$ never changes, they are just grouped into partitions. The used clustering method is purely geometric and cannot adapt to learn semantically meaningful groupings.

\subsection{Graph compression}

Classic compression methods like pruning \cite{Zhou2021AcceleratingLS}, quantization \cite{Zhao2020LearnedLP} or knowledge distillation \cite{Yang2020DistillingKF} have been successfully applied to GNN. 
%Graph Reduction
Graphs can be quite complex and this has led to the exploration of graph condensation to replace wide graphs with smaller versions that are easier to manipulate in term of computations \cite{Hashemi2024ACS}. There are three main categories of graph condensations: graph sparsification, coarsening and condensation.
Graph sparsification can be seen as a form of pruning, it aims to selects existing nodes or edges and remove the rest of the graph.
Graph coarsening is the surjective mapping of a graph that results in a smaller version of the graph made of "super nodes" that aggregate original nodes. 
Finally, graph condensation is the creation of a small synthetic graph that aims to provide similar performances as the original graph. 

There are different graph coarsening techniques. We can cite DiffPool \cite{Ying2018HierarchicalGR}, that learns a soft assignment matrix $\mathbf{S}$ via a full GNN pass, costing $O(N^2)$ memory for the assignment matrix.
% Top-K
The Top-K method \cite{Gao2019GraphU} select a subset of existing nodes and drops the rest, retaining the subgraph induced by selected nodes. This works for sparse graphs but in ViG's dense k-NN setting it creates coverage gaps.
% SAGPool 
SAGPool \cite{Lee2019SelfAttentionGP} is quite similar to Top-K but the score is computed differently using a GCN meaning that this method in contrast to the other is graph-aware.  
% Graclus 
Graclus \cite{Dhillon2007WeightedGC} is a deterministic graph algorithm that matches neighboring nodes. For each node, it selects the node that maximizes a criterion and matches those nodes together. This process can be applied iteratively to coarsen even more.

\section{Method}
\label{sec:preliminaries}

In this section, we will first describe the basic ViG architecture and the MinCUT Pooling algorithm. 
Then, we will introduce our proposed CoViG approach.
%the Vision GNN works, how MinCUT pooling operates, and why using it was the best coarsening option.

\subsection{Vision GNN}

Han \etal \cite{Han2022VisionGA} proposed the first ViG; this method can be considered a replacement for the self-attention mechanism of a Vision Transformer, using graph message passing instead of attention. The image is first divided into visual nodes. Each node corresponds to the receptive field of one feature map location. 

\begin{figure}[ht]
    \centering
    \includegraphics[width=\linewidth]{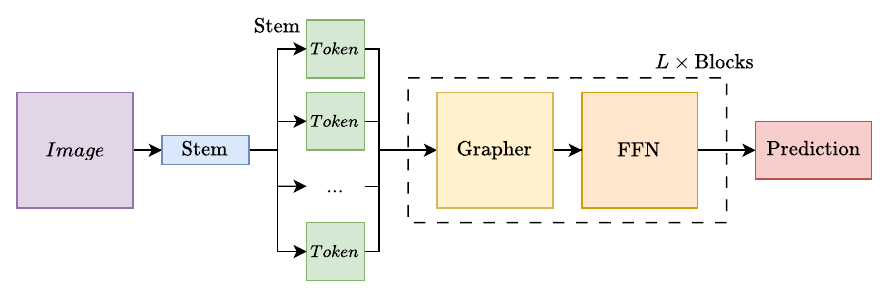}
    \caption{Overview of the Vision Graph Neural Network (ViG) architecture.}
    \label{fig:vig}
\end{figure}

The network is composed of several blocks that repeatedly follow the following path shown in \cref{fig:vig,fig:block}. For each node, ViG computes the $k$ nearest neighbors, producing a graph. This graph is based on feature similarity, not spatial adjacency. It is then passed through a GCN, which gathers neighbour features, aggregates the information and updates the node representation. Each node is then processed by a feed-forward network (FFN). 

% Classification
After the final block, global pooling is applied to all nodes, and the resulting feature vector is fed to a linear classifier.

\begin{figure*}[t]
    \centering
    \begin{subfigure}{.49\textwidth}
        \includegraphics[width=\linewidth]{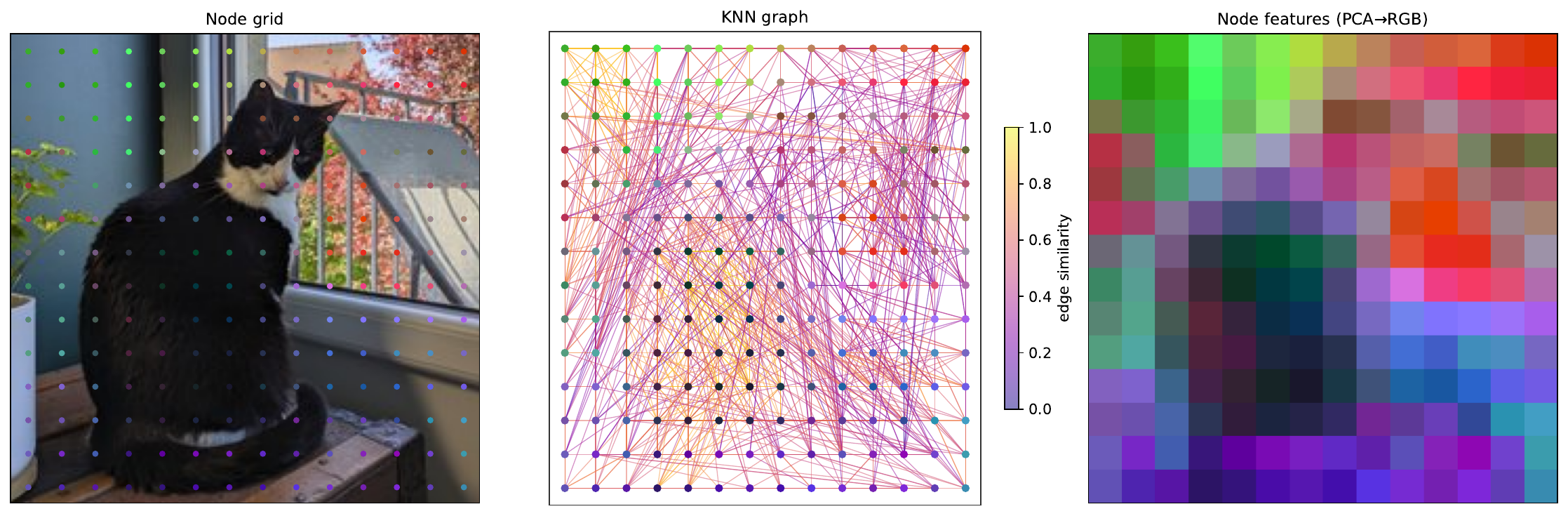}
        \caption{Block 0}
        \label{fig:block_0}
    \end{subfigure}
    \hfill
    \begin{subfigure}{.49\textwidth}
        \includegraphics[width=\linewidth]{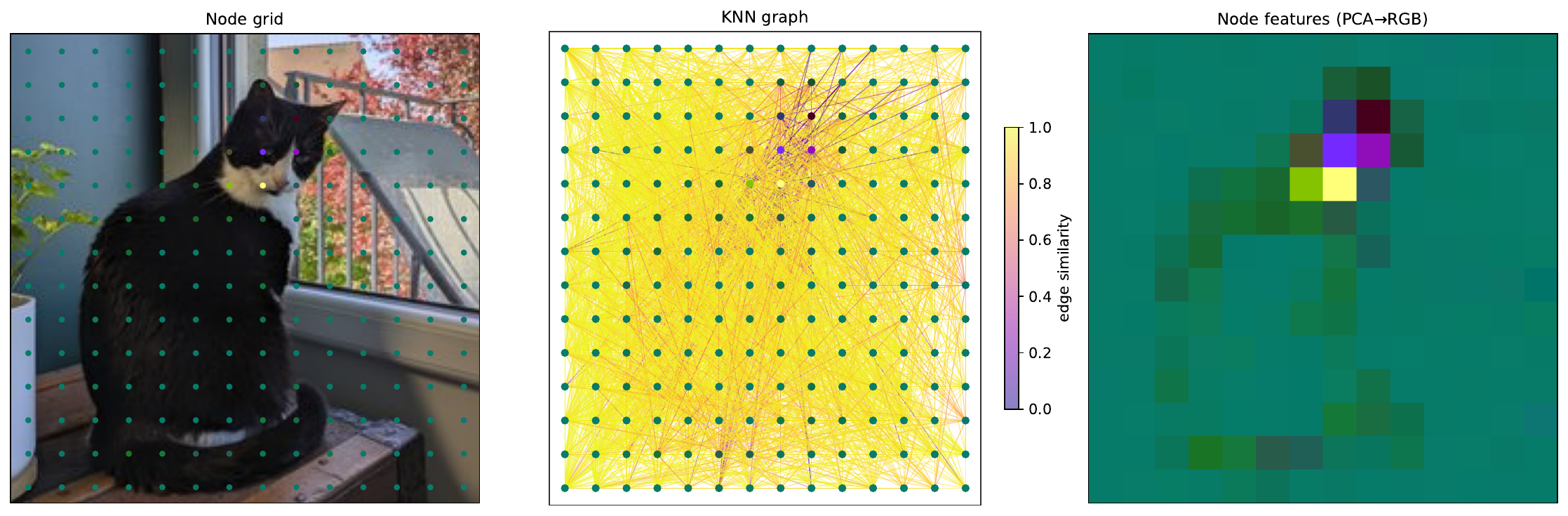}
        \caption{Block 11}
        \label{fig:block_11}
    \end{subfigure}
    \caption{Visualization of the graph representation learned by the ViG \cite{Han2022VisionGA}. From left to right: the input image overlaid with the 14×14 graph nodes, the dynamically constructed k-NN graph based on feature similarity, and a PCA projection of the learned node features into RGB space. Subfigures a) and b) correspond to different backbone blocks, illustrating the evolution of graph connectivity and feature representations throughout the network.}
    \label{fig:vig_block}
\end{figure*}

In terms of computational complexity, consider the ViG-Ti (tiny) architecture: the number of nodes is $N = 196$, the number of feature channels $C = 768$, and the number of neighbors $k = 9$. 
The FLOPs of a single GRAPHER+FFN block scale as:
\begin{equation}
    \underbrace{NC^2}_{fc1} + \underbrace{2N^2C}_{distance} + \underbrace{8NC^2}_{MRConv} + \underbrace{2NC^2}_{fc2}
\end{equation}

\autoref{fig:vig_block} visualizes the k-NN graph structure and node feature representations at two depths of the network: the first block and the last block. 
In the first block, node features exhibit rich spatial diversity and the corresponding k-NN graph is unstructured with edges connecting spatially distant nodes at low similarity scores, encoding meaningful long-range semantic relationships across the image. This confirms that the full node resolution is justified at early depth, where spatial heterogeneity is high and every node contributes discriminative information. 
At the final block, the picture changes, the feature map reveals that the vast majority of nodes have collapsed to a near-uniform representation. Only a handful of patches retains distinct features. Similarly, the k-NN graph degenerates into a uniform high edge similarity graph. 

At the deepest blocks, the expensive computation cost of the Grapher and FFN blocks are not justified because the graph does not carries much structural information. These observations reveal the feature diversity collapse of the ViG and justifies the introduction of graph coarsening to not maintain redundant nodes through the entire network. In our approach, we have chosen to apply a learned pooling operation (MinCUT) that can merge near identical nodes into a set of $K<<N$ super nodes.

\subsection{MinCUT Pooling}

Bianchi \etal \cite{Bianchi2019MincutPI} proposed a pooling / coarsening method that 
produces a soft assignment matrix $\mathbf{S}$ learned via backpropagation. In contrast to ClusterViG \cite{Parikh2025ClusterViGEG} or Graclus \cite{Dhillon2007WeightedGC} that are not differentiable, MinCUT allows model to learn which patch to merge based on the usual task-specific loss - \ie cross entropy (CE) for classification - and the unsupervised loss $\mathcal{L}_u$ (\autoref{eq:loss_u}):
% Loss unsupervised
\begin{equation}
    \mathcal{L}_u = \mathcal{L}_c + \mathcal{L}_o = \underbrace{-\frac{Tr\left(\mathbf{S}^T\mathbf{A}\mathbf{S}\right)}{Tr\left(\mathbf{S}^T\mathbf{D}\mathbf{S}\right)}}_{\mathcal{L}_c} + \underbrace{\left\| \frac{\mathbf{S}^T\mathbf{S}}{\|\mathbf{S}^T\mathbf{S}\|_F} - \frac{\mathbf{I}_K}{\sqrt{K}} \right\|_F}_{\mathcal{L}_o} \; ,
\label{eq:loss_u}
\end{equation}
where $\mathbf{A}$ is the normalized adjacency matrix of the graph, $\mathbf{D} = \text{diag}\left(\mathbf{A}\mathbf{1}_N\right)$ is the degree matrix. $\mathbf{I}_K$ refers to the identity matrix, $\mathbf{1}_N$ refers to the matrix of ones, and $\|.\|_F$, to the Frobenius norm defined as:

% Frobenius norm
\begin{equation}
    \|A\|_F:=\sqrt{Tr(A^*A)}.
\end{equation}

The unsupervised loss is composed of the cut loss $\mathcal{L}_c$ and the orthogonality loss $\mathcal{L}_o$. 

The cut loss explicitly encourages the assignment to respect graph connectivity. It means that strongly connected nodes (nodes sharing similar features, close in feature space) are encouraged to be merged together. This loss prevents degenerate solutions where all nodes collapse to the same cluster. However, when all nodes are assigned to the same cluster, the numerator and denominator are equal and $\mathcal{L}_c$ is minimized.

To avoid such local minimum, $\mathcal{L}_o$ encourages the MLP to find clusters that are orthogonal between each other (\ie avoiding assigning all nodes to all clusters).

Ensuring that each node is assigned to at least one super-node avoids any coverage gaps that can occur with Top-K or SAGPool.

The fact that MinCUT performs soft assignment - contrary to Top-K, SAGPool, Graclus, and ClusterViG - allows patches at boundaries between semantic regions to be assigned to several super-nodes. It preserves boundary information that would be lost by hard methods. 

Comparing to DiffPool, MinCUT computes $\mathbf{S}$ via an MLP on node features costing $O(N \times K)$ memory for the assignment matrix. DiffPool also requires a separate GNN to compute $\mathbf{S}$, it effectively doubles the graph computation at the coarsening block. 

For all these reasons, we have chosen to use MinCUT over other coarsening methods to coarsen ViG.

\begin{comment}
\begin{equation}
    \mathcal{L}_c = -\frac{Tr\left(\mathbf{S}^T\mathbf{A}\mathbf{S}\right)}{Tr\left(\mathbf{S}^T\mathbf{D}\mathbf{S}\right)}
\end{equation}

\begin{equation}
    \mathcal{L}_o = \left\| \frac{\mathbf{S}^T\mathbf{S}}{\|\mathbf{S}^T\mathbf{S}\|_F} - \frac{\mathbf{I}_K}{\sqrt{K}} \right\|_F
\end{equation}
\end{comment}

\begin{comment}
MinCUT produces $\mathbf{A}_{pool} = \mathbf{S}^T \mathbf{A}  \mathbf{S}$
\end{comment}
\subsection{CoViG architecture}

We will now describe how the CoViG architecture integrates MinCUT Pooling in ViG. 

%coarsened Grapher
While the original ViG keeps the same number of nodes through all blocks, CoViG adds a coarsened Grapher to replace the original Grapher. This new Grapher reduces the number of nodes according to different parameters. Several coarsened Grapher can be placed along the model to reduce multiple time the nodes number. The position of the different pooling blocks is defined using two parameters: coarsen start ($C_{start}$) is the index of the first block where coarsening is applied; coarsen step ($C_{step}$) is the distance between two coarsening blocks. 
The coarsening strength (the number of coarsened nodes) is defined by two parameters: coarsening ratio ($C_r$) is the fraction of nodes kept after each coarsening blocks; node threshold ($N_T$) is the minimum number of nodes.

For example, using a ViG-Ti as base model (with 196 nodes and 12 blocks) if the coarsening start on the third block ($C_{start}=3$) with coarsening applied each sixth blocks ($C_{step}=6$), 50\% of the nodes pruned each time ($C_r=0.5$) and a minimum of 49 nodes ($N_T=49$). The final model will apply coarsening on the third block ($196 \rightarrow 98$ nodes) and the ninth block ($98 \rightarrow 49$ nodes). 
 
% MinCUT pool
MinCUT is applied to the graph using the existing k-NN graph and computing the soft assignment matrix $\mathbf{S}$ using a trained MLP. The node features $\mathbf{X}$ is pooled using $\mathbf{X}_{coarse} = \mathbf{XS}$. The graph convolutions are then applied to the newly created $\mathbf{X}_{coarse}$. To train the MinCUT Pooling, the auxiliary loss $\mathcal{L}_u$ is added to the cross-entropy:
\begin{equation}
    \mathcal{L}_{tot} = \mathcal{L}_{\text{CE}} + \omega_u \mathcal{L}_{u}
\end{equation}

The value of $\omega_u$ will be discussed in the following sections. As the number of nodes changed inside the coarsened Grapher, the residual connection from the original Grapher block has been removed in the coarsened version.

\section{Experiments}
\label{sec:experiments}
 
This section provides details on the setup used to perform the experiments.
 
%\subsection{Dataset}
\noindent \textbf{Dataset:} We choose to use ImageNet \cite{Russakovsky2014ImageNetLS} a widely used image classification dataset, for our experiments. It contains 1.2M training images and 50K validation images, distributed across 1000 classes. 

%\subsection{Experimental Settings}
\noindent \textbf{Experimental settings:}
Using the pre-trained models from the original paper as base, we trained the coarsened versions for 150 epochs using a batch size of 128. We kept most of the original parameters, the depth and dimension $D$ remain the same for the 2 models. We keep a $k$ k-NN  in range 9 to 18, the graph convolution used is Max-Relative.
We used a learning rate of $10^{-4}$ with AdamW as an optimizer and cosine (with restarts) as learning rate scheduler.

\noindent \textbf{Metrics:}
% Accuracy 
For all experiments, we report top-1 accuracy as the primary evaluation metric (noted accuracy in the following section). Top-1 accuracy corresponds to the percentage of samples for which the model's highest-confidence prediction matches the ground-truth class. 

% Flops 
The computational cost of a model is calculated using fvcore FlopCountAnalysis. This library provides a computational cost in MACs but since most works \cite{Han2022VisionGA} used 1 FLOPs = 1 MACs, we have chosen to present our results in FLOPs with the same conversion ratio instead of the classic 2 FLOPs = 1 MACs. 

% Latency
The model latency is measured using a batch size of 1. A warm-up of 20 images is used. Then, the latency is averaged over 200 images. The latency is calculated on a single NVIDIA A100 GPU.

\noindent \textbf{Ablations:}
We tested several positions $C_{start}$ for the coarsening block with a fixed coarsening ratio $C_r$ and changed the coarsening ratio while keeping the coarsening block on the same position. 
We also explored the possibility to coarsen multiple times the same model with different gaps between two coarsenings and changing the coarsening rate. 

We also explore the impact of different $\omega_u$ values in range 0 to 2. For all other experiments, we used $\omega_u = 0.1$.

\noindent \textbf{Hardware:} Our experiments have been performed on 3 different hardware settings. To train the CoViG-Ti, we used 8 NVIDIA A100 GPU. For the CoViG-s training, we used 16 NVIDIA A100 GPU on 4 nodes. Finally, we evaluated our models on a single NVIDIA A100 GPU.

\section{Results}
\label{sec:results}

In this section, CoViG-Ti-X-Y model refers to a CoViG-Ti model with a coarsening starting on the X-th block and a compression ratio of Y. 

\begin{table}[ht]
  \centering
  \setlength{\tabcolsep}{0pt}
  \begin{tabular*}{\linewidth}{@{\extracolsep{\fill}}lcccccc@{}}
    \toprule
    Model & $C_{start}$ & $C_r$  & \makecell{Params\\(M)} & \makecell{FLOPs\\(G)} & \makecell{Latency\\(ms)} & \makecell{Acc\\(\%)}\\
    \midrule
    ViG-Ti&- & - & 7.18 & 1.31 & 15.22 & 74.4\\
    \midrule
    \multirow{11}{*}{CoViG-Ti}
    & 6 & 0.5  & 7.21 & 1.03 & 15.19 & \textbf{75.5}\\
    & 3 & 0.5  & 7.21 & 0.88 & 14.64 & 75.0\\
    & 6 & 0.25 & 7.21 & 0.89 & 15.18 & 75.1\\
    & 5 & 0.25 & 7.21 & 0.82 & 15.23 & 74.9\\
    & 4 & 0.25 & 7.21 & 0.74 & 15.08 & 74.9\\
    & 3 & 0.25 & 7.21 & 0.67 & \textbf{14.54} & 74.5\\
    & 2 & 0.25 & 7.21 & 0.60 & 14.87 & 73.8\\
    & 1 & 0.25 & 7.21 & 0.53 & 15.34 & 72.7\\
    & 0 & 0.25 & 7.21 & \textbf{0.45} & 15.31 & 70.3\\
    & 3 & 0.1  & 7.21 & 0.55 & 14.84 & 73.5\\
    & 3 & 0.05 & \textbf{7.20} & 0.51 & 14.82 & 72.5\\
    \bottomrule
  \end{tabular*}
  \caption{CoViG-Ti performances under different single coarsening configurations}
  \label{tab:tiny}
\end{table}

\begin{figure*}[t]
    \centering
    \begin{subfigure}{.495\textwidth}
        \includegraphics[width=\linewidth]{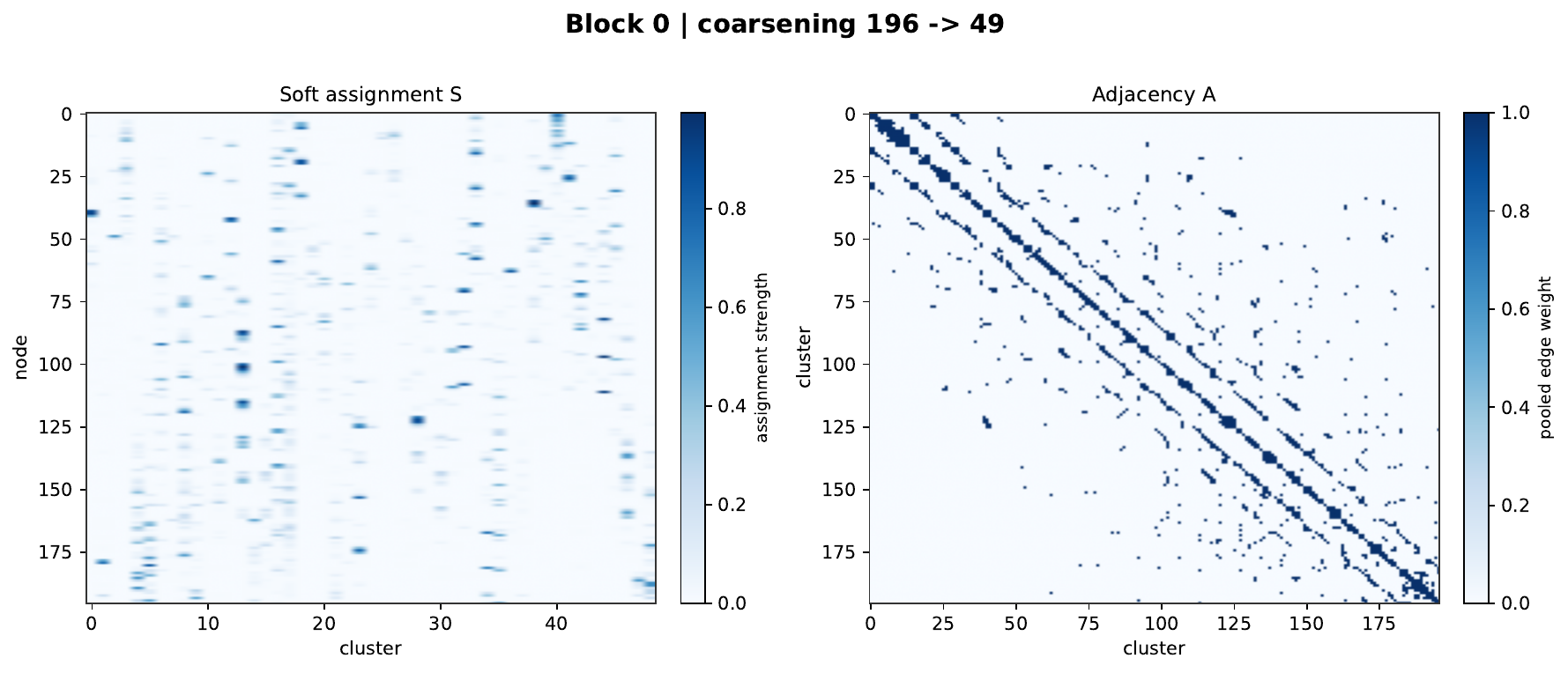}
        \caption{Block 0}
        \label{fig:covig_block_0}
    \end{subfigure}
    \hfill
    \begin{subfigure}{.495\textwidth}
        \includegraphics[width=\linewidth]{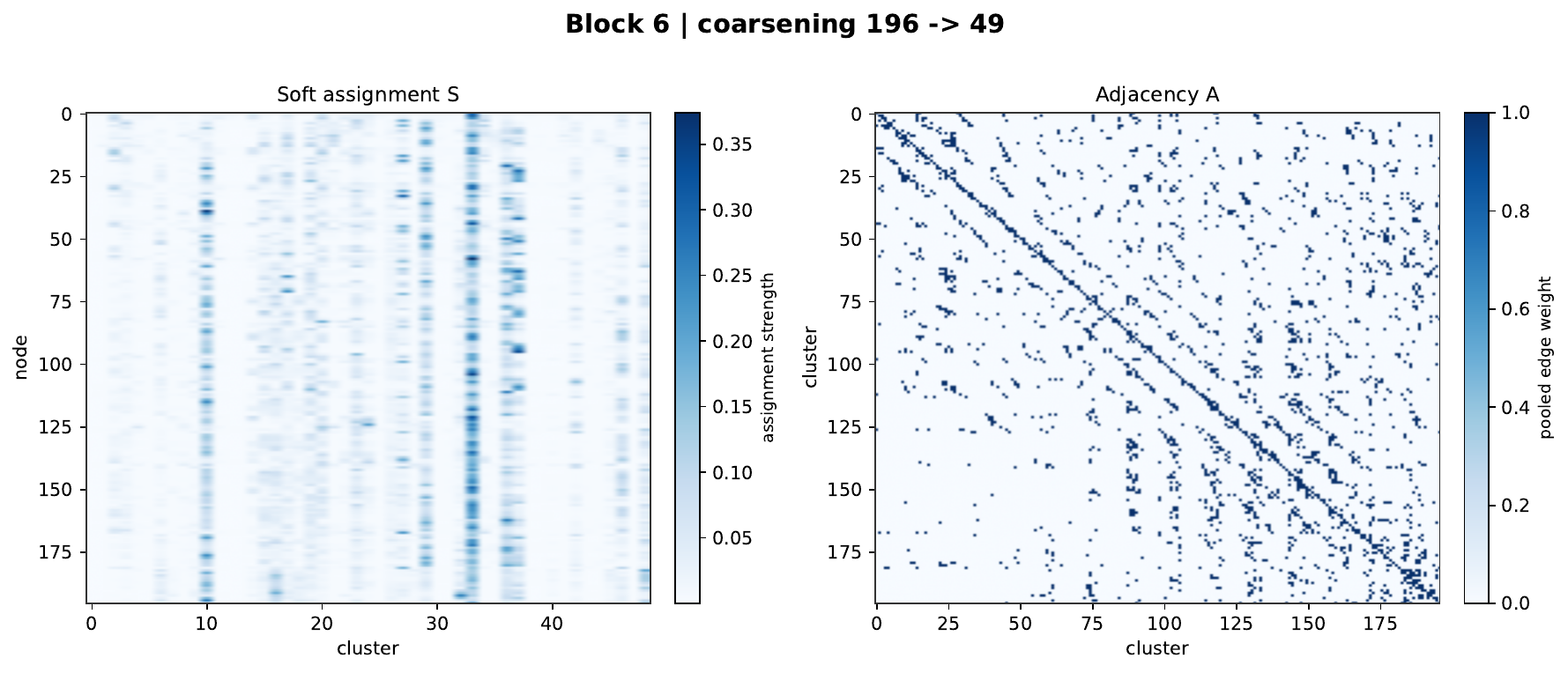}
        \caption{Block 6}
        \label{fig:covig_block_6}
    \end{subfigure}
    \caption{Effect of coarsening depth on MinCUT assignment quality: near-hard spatial assignments at block 0 vs. soft semantic aggregation at block 6.}
    \label{fig:covig_block}
\end{figure*}

We have first tested single coarsening applied on different blocks and with different coarsening ratios on ViG-Ti. These models are presented in \autoref{tab:tiny}. We can note that most of the tested configurations outperform or match the baseline accuracy. For example, the $C_{start} = 6$ and $C_{r} = 0.5$ obtain a one point higher accuracy while only using 78\% of the reference computational cost. $C_{start} = 3$ and $C_{r} = 0.25$ matches the baseline accuracy while reducing the computational cost by half. We can conclude that starting the coarsening after the third block and having a coarsening ratio over 25\% gives more accurate models than the original ViG-Ti. We can also note that the number of parameters increases due to the introduction of a MinCUT Pooling block, but this only adds 0.5\% parameters, which is negligible.

We changed $\omega_u$ to see its impact on the CoViG accuracy. Testing values of 0.1, 1, and 2 we did not find any statistically significant difference in term of accuracy so we kept $\omega_u = 0.1$ for all our experiments.

To qualitatively assess the behavior of the MinCUT Pooling module and validate the choice of coarsening position, we visualize the soft assignment matrix $\mathbf{S} \in \mathcal{R}^{196\times49}$ and the adjacency matrix $\mathbf{A} \in \mathcal{R}^{196\times196}$ for two coarsening configurations: CoViG-0-0.25 and CoViG-6-0.25. These matrices are visualized in \autoref{fig:covig_block}. Several observations emerge from these visualizations: 

First, the adjacency matrices of both configurations exhibit a clear block-diagonal structure, confirming that the MinCUT loss successfully encourages the pooling to preserve intra-cluster connectivity. Clusters are well-separated and super-nodes are connected primarily to other super-nodes within the same semantic group, as theoretically prescribed by the MinCUT objective. However, the adjacency matrices exhibit an important difference: at Block 0, the connectivity is strongly concentrated around the diagonal and a few nearby bands, whereas at Block 6 the connectivity is more dispersed across the graph, with substantially more off-diagonal structure. This indicates that deeper ViG representations contain richer non-local relationships between visual nodes.

Second, the soft assignment matrices reveal a marked qualitative difference between the two configurations. At block 0, assignments are near-hard and high-contrast, with individual nodes assigning almost their entire probability mass to a single cluster with confidence values reaching 0.9. This behavior indicates that early-block features are spatially discriminative enough for the MLP to produce confident, near-deterministic cluster memberships. However, such near-hard assignments provide little benefit over hard partitioning methods and produce super-nodes that aggregate only a handful of original nodes, limiting the semantic richness of the coarsened representation. In contrast, at block 6, assignments are considerably softer and more diffuse, with probability mass distributed across multiple clusters and peak values limited to 0.35. This diffuse structure reflects the feature homogenization that occurs at deeper network layers. As node representations converge toward more uniform embeddings, the assignment MLP naturally distributes contributions across clusters, producing super-nodes that are genuine weighted aggregations of semantically related patches rather than near-singleton partitions. The corresponding pooled adjacency at block 6 is denser and exhibits richer inter-cluster connectivity, indicating that the coarsened graph retains meaningful global structure for the downstream Grapher blocks to operate on. 

To conclude, coarsening at Block 6 can aggregate nodes after the network has established more meaningful global relationships, whereas coarsening at Block 0 operates on a less developed representation. This qualitative observation is consistent with the quantitative results: delaying coarsening generally provides a better accuracy–efficiency trade-off than applying it at the beginning of the network.

\begin{table}[ht]
  \centering
  \setlength{\tabcolsep}{0pt}
  \begin{tabular*}{\linewidth}{@{\extracolsep{\fill}}lccccccc@{}}
    \toprule
    Model & $C_{start}$ & $C_r$  & $C_{step}$ & $N_T$ & \makecell{Params\\(M)} & \makecell{FLOPs\\(G)} & \makecell{Acc\\(\%)}\\
    \midrule
    ViG-Ti &- & - & - & - & 7.18 & 1.31 & 74.4\\
    \midrule
    \multirow{4}{*}{CoViG-Ti}
    & 3 & 0.9 & 1 & 49 & 7.45 & 1.03 &  7.0\\
    & 6 & 0.9 & 1 & 49 & 7.38 & 1.19 & 32.5\\
    & 4 & 0.5 & 4 & 49 & 7.24 & 0.84 & 69.7\\
    & 6 & 0.5 & 3 & 49 & 7.24 & 0.96 & 71.7\\
    \bottomrule
  \end{tabular*}
  \caption{CoViG-Ti performances under different multiple coarsening configurations}
  \label{tab:multi}
\end{table}

\autoref{tab:multi} shows that applying graph coarsening multiple times is less effective than performing a single coarsening operation. While repeated coarsening further reduces the number of visual nodes and consequently the computational cost, it leads to a significant degradation in accuracy. This indicates that successive reductions of the visual graph progressively discard discriminative information. In contrast, a single coarsening stage allows the network to perform graph reasoning at full resolution during the earlier layers before reducing the graph size, providing a better trade-off between computational efficiency and accuracy. For example the first model in \autoref{tab:multi} has the same computational cost as the CoViG-Ti-6-0.5 model but achieves only 7\% of accuracy comparing to the 75.5\% accuracy of the single coarsened model.

\begin{table}[ht]
  \centering
  \setlength{\tabcolsep}{0pt}
  \begin{tabular*}{\linewidth}{@{\extracolsep{\fill}}lccccccc@{}}
    \toprule
    Model & $C_{start}$ & $C_r$  & \makecell{Params\\(M)} & \makecell{FLOPs\\(G)} & \makecell{Latency\\(ms)} & \makecell{Acc\\(\%)}\\
    \midrule
    ViG-S & - & - & 22.7 & 4.54 & 20.31 & 80.6\\
    \midrule
    \multirow{3}{*}{CoViG-S}
    & 8 & 0.5 & 22.81 & 3.52 & \textbf{19.93} & \textbf{78.7}\\
    & 6 & 0.5 & 22.81 & 3.26 & 20.65 & 78.3\\
    & 4 & 0.5 & 22.81 & \textbf{3.0} & 20.37 & 77.8\\
    \bottomrule
  \end{tabular*}
  \caption{CoViG-S performances under different single coarsening configurations}
  \label{tab:small}
\end{table}

The results presented in \autoref{tab:small} show that the trends observed for CoViG-Ti also extend to the larger ViG-S model. Although the accuracy-efficiency trade-off is less favorable. The tested models reduces the FLOPs from 22.5 to 33.9\% while decreasing accuracy by 1.9 to 2.8 accuracy points. These results suggest that graph coarsening is particularly effective for the smaller ViG-Ti model, where a substantial reduction in computational cost can be achieved with limited impact on accuracy, whereas the larger ViG-S model appears more sensitive to the reduction of visual nodes. Nevertheless, CoViG-S still provides a meaningful computational reduction. 

\begin{table}[ht]
  \centering
  \setlength{\tabcolsep}{0pt}
  \begin{tabular*}{\linewidth}{@{\extracolsep{\fill}}lccc@{}}
    \toprule
    Model & \makecell{Params\\(M)} & \makecell{FLOPs\\(G)} & \makecell{Acc\\(\%)}\\
    \midrule
    ViG-Ti \cite{Han2022VisionGA}                
    & 7.18 & 1.31 & 74.4\\
    SVIG-Ti \cite{Elsharkawi2025SViGAS}         
    & 7.2  & 1.3  & 74.6\\
    ViHGNN-Ti \cite{Han2023VisionHA}             
    & 8.2  & 1.8  & 74.3\\
    Isotropic ViDR-GNN-Ti \cite{Zhang2026ViDRGNNVI}
    & 7.3  & 1.4  & 74.6\\
    ViG-LRGC \cite{Elsharkawi2025ViGLRGCVG}
    & 8.1 & 1.5 & 74.3\\
    \midrule
    
    CoViG-Ti-6-0.5 (ours)   
    & 7.21 & \textbf{1.03} & \textbf{75.5}\\
    CoViG-Ti-3-0.25 (ours)          
    & 7.21 & \textbf{0.67} & 74.5\\
    \bottomrule
  \end{tabular*}
  \caption{Comparison of our CoViG method with different isotropic ViG-based models.}
  \label{tab:sota}
\end{table}

We have compared our method against other isotropic ViG based methods (see \autoref{tab:sota}). Our CoViG-Ti-6-0.5 achieves the best accuracy among all models while requiring less computations than the other methods. CoViG-3-0.25 is half the size of the other models while keeping the same accuracy. 
These results demonstrate that graph coarsening can substantially reduce the computational cost of an isotropic ViG architecture while preserving, and in the less aggressive setting even improving, its classification performances.

\section{Conclusion}
\label{sec:conclusion}

In this work, we introduce CoViG, a graph coarsening approach to reduce the computational cost of isotropic Vision GNNs while preserving their classification performances. CoViG integrates MinCUT Pooling into the ViG architecture to progressively reduce the number of visual tokens processed by subsequent graph convolutional blocks. This reduces the computational cost of graph reasoning while keeping the original depth and feature dimensionality of the ViG architecture.

Our experiments on ImageNet demonstrate that the effectiveness of coarsening strongly depends on the coarsened Grapher position in the network. Applying coarsening too early can discard spatial information before sufficiently high-level representations have been formed, whereas delaying coarsening allows the network to perform graph reasoning at full resolution before aggregating visual tokens. In particular, for ViG-Ti, CoViG can reduce the computational cost by approximately half while maintaining the baseline accuracy, and a less aggressive configuration even improves the accuracy from 74.4\% to 75.5\% while reducing FLOPs from 1.31G to 1.03G. These results indicate that graph coarsening can provide computational savings without necessarily requiring a proportional loss in recognition performance.

We further showed that repeatedly coarsening the graph is less effective than applying a single coarsening operation at an appropriate depth. Although multiple coarsening stages can further reduce the number of visual nodes, they can progressively discard discriminative information and result in substantial accuracy degradation. Experiments with the larger ViG-S model confirm the generality of the approach, while also showing that larger models can be more sensitive to aggressive node reduction.

Overall, these results demonstrate that visual tokens in graph-based vision architectures do not necessarily need to be processed at full resolution throughout the entire network. Carefully positioned graph coarsening provides a simple mechanism for controlling the computational cost of graph reasoning while preserving, and in some configurations improving, recognition accuracy. Future work could investigate adaptive coarsening strategies that determine the number of visual nodes dynamically based on the input or the semantic content of intermediate representations, as well as extending CoViG to other graph-based vision architectures (\ie pyramid ViG).

{
    \small
    \bibliographystyle{ieeenat_fullname}
    \bibliography{main}

@String(CVPR= {IEEE Conf. Comput. Vis. Pattern Recog.})

@String(ICCV= {Int. Conf. Comput. Vis.})

@String(ICPR = {Int. Conf. Pattern Recog.})

@String(TOG= {ACM Trans. Graph.})

@String(AAAI = {AAAI})

@String(CVPRW= {IEEE Conf. Comput. Vis. Pattern Recog. Worksh.})

@String(CVPR  = {CVPR})

@String(ICCV  = {ICCV})

@String(ICPR  = {ICPR})

@String(TOG   = {ACM TOG})

@String(CVPRW= {CVPRW})

@article{LeCun1998GradientbasedLA,
  title={Gradient-based learning applied to document recognition},
  author={Yann LeCun and L{\'e}on Bottou and Yoshua Bengio and Patrick Haffner},
  journal={Proc. IEEE},
  year={1998},
  volume={86},
  pages={2278-2324},
  url={https://api.semanticscholar.org/CorpusID:14542261}
}

@article{Krizhevsky2012ImageNetCW,
  title={ImageNet classification with deep convolutional neural networks},
  author={Alex Krizhevsky and Ilya Sutskever and Geoffrey E. Hinton},
  journal={Communications of the ACM},
  year={2012},
  volume={60},
  pages={84 - 90},
  url={https://api.semanticscholar.org/CorpusID:195908774}
}

@article{Ren2015FasterRT,
  title={Faster R-CNN: Towards Real-Time Object Detection with Region Proposal Networks},
  author={Shaoqing Ren and Kaiming He and Ross B. Girshick and Jian Sun},
  journal={IEEE Transactions on Pattern Analysis and Machine Intelligence},
  year={2015},
  volume={39},
  pages={1137-1149},
  url={https://api.semanticscholar.org/CorpusID:10328909}
}

@article{Shelhamer2014FullyCN,
  title={Fully convolutional networks for semantic segmentation},
  author={Evan Shelhamer and Jonathan Long and Trevor Darrell},
  journal={2015 IEEE Conference on Computer Vision and Pattern Recognition (CVPR)},
  year={2014},
  pages={3431-3440},
  url={https://api.semanticscholar.org/CorpusID:1629541}
}

@article{Simonyan2014VeryDC,
  title={Very Deep Convolutional Networks for Large-Scale Image Recognition},
  author={Karen Simonyan and Andrew Zisserman},
  journal={CoRR},
  year={2014},
  volume={abs/1409.1556},
  url={https://api.semanticscholar.org/CorpusID:14124313}
}

@article{He2015DeepRL,
  title={Deep Residual Learning for Image Recognition},
  author={Kaiming He and X. Zhang and Shaoqing Ren and Jian Sun},
  journal={2016 IEEE Conference on Computer Vision and Pattern Recognition (CVPR)},
  year={2015},
  pages={770-778},
  url={https://api.semanticscholar.org/CorpusID:206594692}
}

@article{Howard2017MobileNetsEC,
  title={MobileNets: Efficient Convolutional Neural Networks for Mobile Vision Applications},
  author={Andrew G. Howard and Menglong Zhu and Bo Chen and Dmitry Kalenichenko and Weijun Wang and Tobias Weyand and Marco Andreetto and Hartwig Adam},
  journal={ArXiv},
  year={2017},
  volume={abs/1704.04861},
  url={https://api.semanticscholar.org/CorpusID:12670695}
}

@inproceedings{Han2015LearningBW,
  title={Learning both Weights and Connections for Efficient Neural Network},
  author={Song Han and Jeff Pool and John Tran and William J. Dally},
  booktitle={Neural Information Processing Systems},
  year={2015},
  url={https://api.semanticscholar.org/CorpusID:2238772}
}

@inproceedings{Srinivas2015DatafreePP,
  title={Data-free Parameter Pruning for Deep Neural Networks},
  author={Suraj Srinivas and R. Venkatesh Babu},
  booktitle={British Machine Vision Conference},
  year={2015},
  url={https://api.semanticscholar.org/CorpusID:15647317}
}

@article{Gong2014CompressingDC,
  title={Compressing Deep Convolutional Networks using Vector Quantization},
  author={Yunchao Gong and L. Liu and Ming Yang and Lubomir D. Bourdev},
  journal={ArXiv},
  year={2014},
  volume={abs/1412.6115},
  url={https://api.semanticscholar.org/CorpusID:6251653}
}

@inproceedings{Gupta2015DeepLW,
  title={Deep Learning with Limited Numerical Precision},
  author={Suyog Gupta and Ankur Agrawal and K. Gopalakrishnan and Pritish Narayanan},
  booktitle={International Conference on Machine Learning},
  year={2015},
  url={https://api.semanticscholar.org/CorpusID:2547043}
}

@article{Cheng2017ASO,
  title={A Survey of Model Compression and Acceleration for Deep Neural Networks},
  author={Yu Cheng and Duo Wang and Pan Zhou and Zhang Tao},
  journal={ArXiv},
  year={2017},
  volume={abs/1710.09282},
  url={https://api.semanticscholar.org/CorpusID:22163846}
}

@article{Han2021DynamicNN,
  title={Dynamic Neural Networks: A Survey},
  author={Yizeng Han and Gao Huang and Shiji Song and Le Yang and Honghui Wang and Yulin Wang},
  journal={IEEE Transactions on Pattern Analysis and Machine Intelligence},
  year={2021},
  volume={44},
  pages={7436-7456},
  url={https://api.semanticscholar.org/CorpusID:231855426}
}

@article{Park2015BiglittleDN,
  title={Big/little deep neural network for ultra low power inference},
  author={Eunhyeok Park and Dongyoung Kim and Soobeom Kim and Yong-Deok Kim and Gunhee Kim and Sungroh Yoon and Sungjoo Yoo},
  journal={2015 International Conference on Hardware/Software Codesign and System Synthesis (CODES+ISSS)},
  year={2015},
  pages={124-132},
  url={https://api.semanticscholar.org/CorpusID:265339}
}

@article{Teerapittayanon2016BranchyNetFI,
  title={BranchyNet: Fast inference via early exiting from deep neural networks},
  author={Surat Teerapittayanon and Bradley McDanel and H. T. Kung},
  journal={2016 23rd International Conference on Pattern Recognition (ICPR)},
  year={2016},
  pages={2464-2469},
  url={https://api.semanticscholar.org/CorpusID:2916466}
}

@inproceedings{Ba2013DoDN,
  title={Do Deep Nets Really Need to be Deep?},
  author={Jimmy Ba and Rich Caruana},
  booktitle={Neural Information Processing Systems},
  year={2013},
  url={https://api.semanticscholar.org/CorpusID:11536917}
}

@article{Hinton2015DistillingTK,
  title={Distilling the Knowledge in a Neural Network},
  author={Geoffrey E. Hinton and Oriol Vinyals and Jeffrey Dean},
  journal={ArXiv},
  year={2015},
  volume={abs/1503.02531},
  url={https://api.semanticscholar.org/CorpusID:7200347}
}

@article{Romero2014FitNetsHF,
  title={FitNets: Hints for Thin Deep Nets},
  author={Adriana Romero and Nicolas Ballas and Samira Ebrahimi Kahou and Antoine Chassang and Carlo Gatta and Yoshua Bengio},
  journal={CoRR},
  year={2014},
  volume={abs/1412.6550},
  url={https://api.semanticscholar.org/CorpusID:2723173}
}

@inproceedings{Vaswani2017AttentionIA,
  title={Attention is All you Need},
  author={Ashish Vaswani and Noam Shazeer and Niki Parmar and Jakob Uszkoreit and Llion Jones and Aidan N. Gomez and Lukasz Kaiser and Illia Polosukhin},
  booktitle={Neural Information Processing Systems},
  year={2017},
  url={https://api.semanticscholar.org/CorpusID:13756489}
}

@article{Dosovitskiy2020AnII,
  title={An Image is Worth 16x16 Words: Transformers for Image Recognition at Scale},
  author={Alexey Dosovitskiy and Lucas Beyer and Alexander Kolesnikov and Dirk Weissenborn and Xiaohua Zhai and Thomas Unterthiner and Mostafa Dehghani and Matthias Minderer and Georg Heigold and Sylvain Gelly and Jakob Uszkoreit and Neil Houlsby},
  journal={ArXiv},
  year={2020},
  volume={abs/2010.11929},
  url={https://api.semanticscholar.org/CorpusID:225039882}
}

@article{Wang2021PyramidVT,
  title={Pyramid Vision Transformer: A Versatile Backbone for Dense Prediction without Convolutions},
  author={Wenhai Wang and Enze Xie and Xiang Li and Deng-Ping Fan and Kaitao Song and Ding Liang and Tong Lu and Ping Luo and Ling Shao},
  journal={2021 IEEE/CVF International Conference on Computer Vision (ICCV)},
  year={2021},
  pages={548-558},
  url={https://api.semanticscholar.org/CorpusID:232035922}
}

@article{Liu2021SwinTH,
  title={Swin Transformer: Hierarchical Vision Transformer using Shifted Windows},
  author={Ze Liu and Yutong Lin and Yue Cao and Han Hu and Yixuan Wei and Zheng Zhang and Stephen Lin and Baining Guo},
  journal={2021 IEEE/CVF International Conference on Computer Vision (ICCV)},
  year={2021},
  pages={9992-10002},
  url={https://api.semanticscholar.org/CorpusID:232352874}
}

@inproceedings{Touvron2020TrainingDI,
  title={Training data-efficient image transformers \& distillation through attention},
  author={Hugo Touvron and Matthieu Cord and Matthijs Douze and Francisco Massa and Alexandre Sablayrolles and Herv{\'e} J{\'e}gou},
  booktitle={International Conference on Machine Learning},
  year={2020},
  url={https://api.semanticscholar.org/CorpusID:229363322}
}

@article{Liu2021PostTrainingQF,
  title={Post-Training Quantization for Vision Transformer},
  author={Zhenhua Liu and Yunhe Wang and Kai Han and Siwei Ma and Wen Gao},
  journal={ArXiv},
  year={2021},
  volume={abs/2106.14156},
  url={https://api.semanticscholar.org/CorpusID:235658553}
}

@article{Rao2021DynamicViTEV,
  title={DynamicViT: Efficient Vision Transformers with Dynamic Token Sparsification},
  author={Yongming Rao and Wenliang Zhao and Benlin Liu and Jiwen Lu and Jie Zhou and Cho-Jui Hsieh},
  journal={ArXiv},
  year={2021},
  volume={abs/2106.02034},
  url={https://api.semanticscholar.org/CorpusID:235313562}
}

@article{Bolya2022TokenMY,
  title={Token Merging: Your ViT But Faster},
  author={Daniel Bolya and Cheng-Yang Fu and Xiaoliang Dai and Peizhao Zhang and Christoph Feichtenhofer and Judy Hoffman},
  journal={ArXiv},
  year={2022},
  volume={abs/2210.09461},
  url={https://api.semanticscholar.org/CorpusID:252968113}
}

@article{Micheli2009NeuralNF,
  title={Neural Network for Graphs: A Contextual Constructive Approach},
  author={Alessio Micheli},
  journal={IEEE Transactions on Neural Networks},
  year={2009},
  volume={20},
  pages={498-511},
  url={https://api.semanticscholar.org/CorpusID:17486263}
}

@article{Gori2005ANM,
  title={A new model for learning in graph domains},
  author={Marco Gori and Gabriele Monfardini and Franco Scarselli},
  journal={Proceedings. 2005 IEEE International Joint Conference on Neural Networks, 2005.},
  year={2005},
  volume={2},
  pages={729-734 vol. 2},
  url={https://api.semanticscholar.org/CorpusID:275758328}
}

@article{Duvenaud2015ConvolutionalNO,
  title={Convolutional Networks on Graphs for Learning Molecular Fingerprints},
  author={David Kristjanson Duvenaud and Dougal Maclaurin and Jorge Aguilera-Iparraguirre and Rafael G{\'o}mez-Bombarelli and Timothy D. Hirzel and Al{\'a}n Aspuru-Guzik and Ryan P. Adams},
  journal={ArXiv},
  year={2015},
  volume={abs/1509.09292},
  url={https://api.semanticscholar.org/CorpusID:1690180}
}

@article{Landrieu2017LargeScalePC,
  title={Large-Scale Point Cloud Semantic Segmentation with Superpoint Graphs},
  author={Loic Landrieu and Martin Simonovsky},
  journal={2018 IEEE/CVF Conference on Computer Vision and Pattern Recognition},
  year={2017},
  pages={4558-4567},
  url={https://api.semanticscholar.org/CorpusID:4396837}
}

@article{Wang2018DynamicGC,
  title={Dynamic Graph CNN for Learning on Point Clouds},
  author={Yue Wang and Yongbin Sun and Ziwei Liu and Sanjay E. Sarma and Michael M. Bronstein and Justin M. Solomon},
  journal={ACM Transactions on Graphics (TOG)},
  year={2018},
  volume={38},
  pages={1 - 12},
  url={https://api.semanticscholar.org/CorpusID:94822}
}

@article{Xu2017SceneGG,
  title={Scene Graph Generation by Iterative Message Passing},
  author={Danfei Xu and Yuke Zhu and Christopher Bongsoo Choy and Li Fei-Fei},
  journal={2017 IEEE Conference on Computer Vision and Pattern Recognition (CVPR)},
  year={2017},
  pages={3097-3106},
  url={https://api.semanticscholar.org/CorpusID:1780254}
}

@inproceedings{Yan2018SpatialTG,
  title={Spatial Temporal Graph Convolutional Networks for Skeleton-Based Action Recognition},
  author={Sijie Yan and Yuanjun Xiong and Dahua Lin},
  booktitle={AAAI Conference on Artificial Intelligence},
  year={2018},
  url={https://api.semanticscholar.org/CorpusID:19167105}
}

@article{Han2022VisionGA,
  title={{Vision GNN: An Image is Worth Graph of Nodes}},
  author={Kai Han and Yunhe Wang and Jianyuan Guo and Yehui Tang and Enhua Wu},
  journal={ArXiv},
  year={2022},
  volume={abs/2206.00272},
  url={https://api.semanticscholar.org/CorpusID:249240094}
}

@article{Munir2023MobileViGGS,
  title={MobileViG: Graph-Based Sparse Attention for Mobile Vision Applications},
  author={Mustafa Munir and William Avery and Radu Marculescu},
  journal={2023 IEEE/CVF Conference on Computer Vision and Pattern Recognition Workshops (CVPRW)},
  year={2023},
  pages={2211-2219},
  url={https://api.semanticscholar.org/CorpusID:259317049}
}

@article{Munir2024GreedyViGDA,
  title={GreedyViG: Dynamic Axial Graph Construction for Efficient Vision GNNs},
  author={Mustafa Munir and William Avery and Md Mostafijur Rahman and Radu Marculescu},
  journal={2024 IEEE/CVF Conference on Computer Vision and Pattern Recognition (CVPR)},
  year={2024},
  pages={6118-6127},
  url={https://api.semanticscholar.org/CorpusID:269757862}
}

@article{Parikh2025ClusterViGEG,
  title={ClusterViG: Efficient Globally Aware Vision GNNs via Image Partitioning},
  author={Dhruv Parikh and Jacob Fein-Ashley and Tian Ye and Rajgopal Kannan and Viktor K. Prasanna},
  journal={2025 International Conference on Multimedia Computing, Networking and Applications (MCNA)},
  year={2025},
  pages={136-143},
  url={https://api.semanticscholar.org/CorpusID:275758269}
}

@article{Zhou2021AcceleratingLS,
  title={Accelerating Large Scale Real-Time GNN Inference using Channel Pruning},
  author={Hongkuan Zhou and Ajitesh Srivastava and Hanqing Zeng and Rajgopal Kannan and Viktor K. Prasanna},
  journal={ArXiv},
  year={2021},
  volume={abs/2105.04528},
  url={https://api.semanticscholar.org/CorpusID:234343119}
}

@article{Zhao2020LearnedLP,
  title={Learned Low Precision Graph Neural Networks},
  author={Yiren Zhao and Duo Wang and Daniel Bates and Robert D. Mullins and Mateja Jamnik and Pietro Li{\`o}},
  journal={ArXiv},
  year={2020},
  volume={abs/2009.09232},
  url={https://api.semanticscholar.org/CorpusID:221818866}
}

@article{Yang2020DistillingKF,
  title={Distilling Knowledge From Graph Convolutional Networks},
  author={Yiding Yang and Jiayan Qiu and Mingli Song and Dacheng Tao and Xinchao Wang},
  journal={2020 IEEE/CVF Conference on Computer Vision and Pattern Recognition (CVPR)},
  year={2020},
  pages={7072-7081},
  url={https://api.semanticscholar.org/CorpusID:214713848}
}

@inproceedings{Hashemi2024ACS,
  title={A Comprehensive Survey on Graph Reduction: Sparsification, Coarsening, and Condensation},
  author={Mohammad Hashemi and Shengbo Gong and Juntong Ni and Wenqi Fan and B. Aditya Prakash and Wei Jin},
  booktitle={International Joint Conference on Artificial Intelligence},
  year={2024},
  url={https://api.semanticscholar.org/CorpusID:267341687}
}

@article{Bianchi2019MincutPI,
  title={Mincut pooling in Graph Neural Networks},
  author={Filippo Maria Bianchi and Daniele Grattarola and Cesare Alippi},
  journal={ArXiv},
  year={2019},
  volume={abs/1907.00481},
  url={https://api.semanticscholar.org/CorpusID:195766901}
}

@inproceedings{Ying2018HierarchicalGR,
  title={Hierarchical Graph Representation Learning with Differentiable Pooling},
  author={Rex Ying and Jiaxuan You and Christopher Morris and Xiang Ren and William L. Hamilton and Jure Leskovec},
  booktitle={Neural Information Processing Systems},
  year={2018},
  url={https://api.semanticscholar.org/CorpusID:49420315}
}

@article{Gao2019GraphU,
  title={Graph U-Nets},
  author={Hongyang Gao and Shuiwang Ji},
  journal={IEEE Transactions on Pattern Analysis and Machine Intelligence},
  year={2019},
  volume={44},
  pages={4948-4960},
  url={https://api.semanticscholar.org/CorpusID:153311899}
}

@article{Dhillon2007WeightedGC,
  title={Weighted Graph Cuts without Eigenvectors A Multilevel Approach},
  author={Inderjit S. Dhillon and Yuqiang Guan and Brian Kulis},
  journal={IEEE Transactions on Pattern Analysis and Machine Intelligence},
  year={2007},
  volume={29},
  url={https://api.semanticscholar.org/CorpusID:9402790}
}

@article{Lee2019SelfAttentionGP,
  title={Self-Attention Graph Pooling},
  author={Junhyun Lee and Inyeop Lee and Jaewoo Kang},
  journal={ArXiv},
  year={2019},
  volume={abs/1904.08082},
  url={https://api.semanticscholar.org/CorpusID:119314157}
}

@article{Russakovsky2014ImageNetLS,
  title={ImageNet Large Scale Visual Recognition Challenge},
  author={Olga Russakovsky and Jia Deng and Hao Su and Jonathan Krause and Sanjeev Satheesh and Sean Ma and Zhiheng Huang and Andrej Karpathy and Aditya Khosla and Michael S. Bernstein and Alexander C. Berg and Li Fei-Fei},
  journal={International Journal of Computer Vision},
  year={2014},
  volume={115},
  pages={211 - 252},
  url={https://api.semanticscholar.org/CorpusID:2930547}
}

@article{Han2023VisionHA,
  title={Vision HGNN: An Image is More than a Graph of Nodes},
  author={Yan Han and Peihao Wang and Souvik Kundu and Ying Ding and Zhangyang Wang},
  journal={2023 IEEE/CVF International Conference on Computer Vision (ICCV)},
  year={2023},
  pages={19821-19831},
  url={https://api.semanticscholar.org/CorpusID:267024787}
}

@article{Zhang2026ViDRGNNVI,
  title={ViDR-GNN: Vision Implicit Discriminative Reorganization Graph Neural Networks},
  author={Zeyang Zhang and Xiao Cao and Xiang Zhang and Li Peng and Lijia Ma and Jielong Yang},
  journal={IEEE Transactions on Multimedia},
  year={2026},
  volume={28},
  pages={2172-2185},
  url={https://api.semanticscholar.org/CorpusID:283587909}
}

@article{Elsharkawi2025SViGAS,
  title={SViG: A Similarity-Thresholded Approach for Vision Graph Neural Networks},
  author={Ismael Elsharkawi and Hossam Sharara and Ahmed A. Rafea},
  journal={IEEE Access},
  year={2025},
  volume={13},
  pages={19379-19387},
  url={https://api.semanticscholar.org/CorpusID:275778795}
}

@article{Elsharkawi2025ViGLRGCVG,
  title={ViG-LRGC: Vision Graph Neural Networks with Learnable Reparameterized Graph Construction},
  author={Ismael Elsharkawi and Hossam Sharara and Ahmed A. Rafea},
  journal={Pattern Recognit. Lett.},
  year={2025},
  volume={207},
  pages={36-41},
  url={https://api.semanticscholar.org/CorpusID:281496591}
}
}

\end{document}